\documentclass[conference]{IEEEtran}
\IEEEoverridecommandlockouts
\usepackage{cite}
\usepackage{amsmath,amssymb,amsfonts}
\usepackage{graphicx}
\usepackage{xcolor}
\usepackage{tikz}
\usetikzlibrary{positioning,calc}
\usepackage{textcomp}
\usepackage{booktabs}
\usepackage{siunitx}
\usepackage[hidelinks]{hyperref}

\graphicspath{{figures/}}

\DeclareRobustCommand{\rev}[1]{#1}

\newif\ifanonymous
\anonymousfalse

\begin{document}

\title{Learning a Whole-Body Pivot-Walking Gait\\
for Heavy-Box Transport with a Humanoid}

\ifanonymous
  \hypersetup{pdfauthor={}}
  \author{\IEEEauthorblockN{Anonymous Authors}}
\else
  \hypersetup{pdfauthor={Rikuto Hidaka, Tsung-Yuan Chen, and Kento Kawaharazuka},
    pdftitle={Learning a Whole-Body Pivot-Walking Gait for Heavy-Box Transport with a Humanoid}}
  \author{Rikuto Hidaka$^{1}$, Tsung-Yuan Chen$^{2}$ and Kento Kawaharazuka$^{3}$%
    \thanks{\scriptsize\raggedright $^{1}$College of Arts and Sciences, The University
      of Tokyo, Tokyo, Japan. \texttt{rhdkreal@gmail.com}}%
    \thanks{\scriptsize\raggedright $^{2}$Next Generation Artificial Intelligence
      Research Center, Graduate School of Information Science and Technology,
      The University of Tokyo, Tokyo, Japan. \texttt{chen.tsungyuan@g.ecc.u-tokyo.ac.jp}}%
    \thanks{\scriptsize\raggedright $^{3}$Department of Mechano-Informatics, Graduate
      School of Information Science and Technology, The University of Tokyo,
      Tokyo, Japan. \texttt{kawaharazuka@jsk.imi.i.u-tokyo.ac.jp}}%
  }
\fi

\maketitle

\begin{abstract}
Through pivot walking, humans can transport objects that are too
heavy to lift.
Previous humanoid studies have relied on planned sequences
in which object manipulation and stepping are executed separately and the 
object is brought to rest between successive rotations. However, 
few studies have developed a single learned controller for humanoid pivot walking.
In this paper, we propose a single reinforcement-learning policy that jointly 
coordinates stepping, balance control, and object manipulation. The policy is 
trained using rewards that promote alternating pivot motions and a phase 
reference that progresses according to the displacement of the object. 
In simulation, we confirmed that the humanoid
can pivot a box. \rev{For deployment, we fine-tuned the policy into an
object-blind version that follows a periodic clock instead of the
progress reference. On the real robot, the deployed policy} transported
the box forward by more than $\SI{1}{m}$ in all five trials\rev{; each
trial ended at a base-tilt safety stop}.
These results demonstrate that coordinated humanoid pivot walking can be
learned with reinforcement learning and transferred to a physical
humanoid robot without direct object-state sensing.
 \end{abstract}

\section{Introduction}
\label{sec:introduction}

\begin{figure}[t]
  \centering
  \includegraphics[width=\columnwidth]{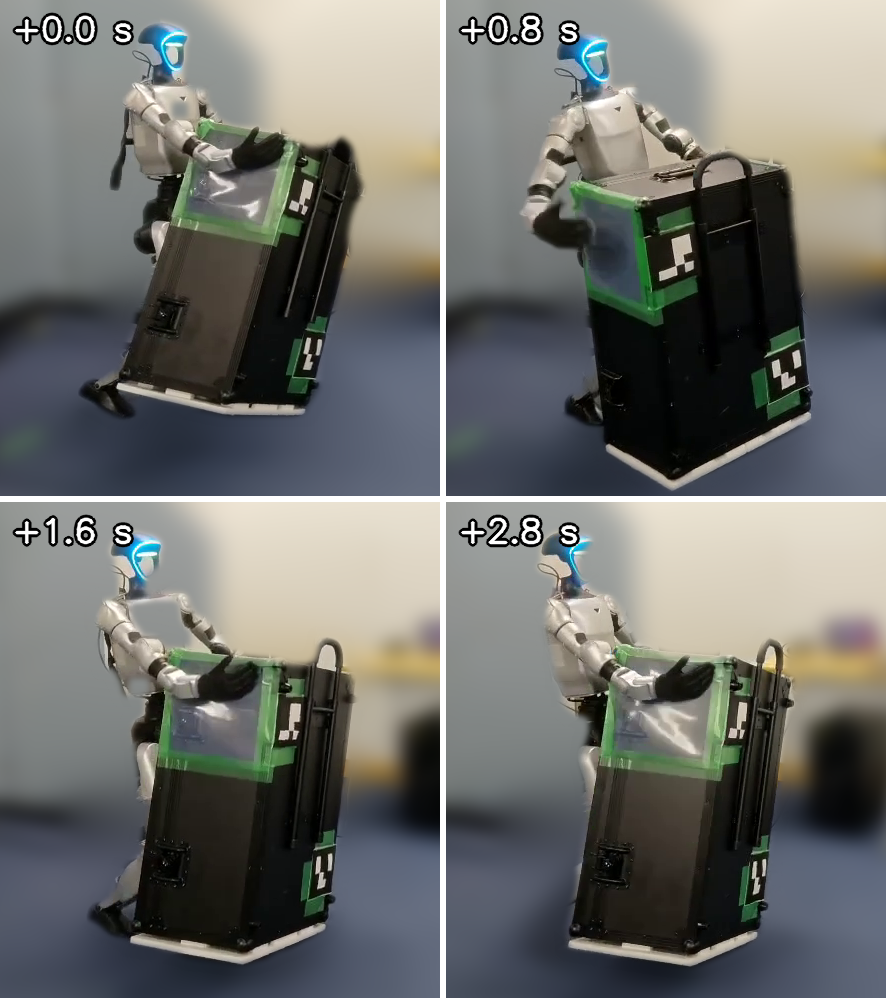}
  \caption{\textbf{Pivot walking experiment.} A Unitree G1 humanoid
  transports a $\SI{14.9}{kg}$ box by tilting it onto a bottom corner
  and rotating it about alternating corners. Representative frames from
  one alternating lean cycle in a hardware trial\rev{. Background
  blurred.}}
  \label{fig:hardware_oblique}
\end{figure}

Humanoid robots are becoming increasingly capable of dynamic locomotion,
yet many useful tasks require moving objects that are too large or heavy
to lift. Direct lifting may exceed the robot's arm capacity, while pushing
or dragging can require large sustained forces and substantial sliding at
the floor. Humans can instead exploit ground support through pivot
walking: the object is rocked between its bottom corners and rotated
forward while most of its weight remains supported by the ground. Pivot
walking may therefore provide humanoids with a useful transport mode for
heavy objects (Fig.~\ref{fig:hardware_oblique}).

Pivoting has been studied extensively as a nonprehensile manipulation primitive,
including for humanoid transport (Section~\ref{sec:related_work}). Existing
humanoid systems, however, rely on explicit contact models and planned
quasi-static sequences. Object rotation and robot stepping are scheduled as
separate phases, and the object is nearly stationary between rotations.
These methods demonstrate reliable task-level planning, yet they do not
realize pivot walking as a continuously executed learned gait in which
stepping, balance, and object rocking are generated jointly.

In this work, we propose such a whole-body pivot-walking gait with
reinforcement learning. In the \rev{six separately trained} routes we
tested, a basic transport objective produces tilted dragging or unstable
pushing rather than repeated pivoting. We introduce box-level rewards that encourage
alternating rocking and a spatial reference indexed by realized box
progress. This reference remains synchronized with the object when
motion slows or stalls. We then transfer the learned gait to a
proprioceptive policy by removing box-state observations and continuing
reinforcement learning with clock conditioning
(Section~\ref{sec:method_blind}). \rev{Progress pacing is thus used
during training, while the deployed policy follows an internal clock.}

In simulation, $\pi_{\mathrm{path}}$
transports a $\SI{14.9}{kg}$ box $\SI{4.12}{m}$ in $\SI{20}{s}$ with
zero observed terminations across 128 \rev{initial box orientations}
(Section~\ref{sec:experiments}). In real robot experiments, the object-blind policy reproduces repeated lean and yaw
reversals and achieves more than $\SI{1}{m}$ of forward transport in all
five trials. All trials terminate at the robot's base-tilt safety
trigger, so the present result demonstrates repeatable short-horizon
gait transfer rather than sustained autonomous delivery
(Section~\ref{sec:exp_real}).

Our contributions are:
\begin{itemize}
  \item A whole-body humanoid policy that walks, balances, and pivots
        a free-standing box with both hands in a single gait, learned
        without an explicit object-contact model or a planned settling
        phase.
  \item \rev{Box-level pivoting rewards and a progress-paced
        reference with which the alternating gait is learned; in our
        tests, a transport reward alone led to dragging or unstable
        pushing.}
  \item Repeatable short-horizon transfer to the physical G1 without
        box-state sensing, achieving repeated lean and yaw reversals
        and transporting the box forward by more than $\SI{1}{m}$ in
        all five trials.
\end{itemize}
\section{Related Work}
\label{sec:related_work}

\subsection{Reinforcement Learning for Legged Locomotion}
Reinforcement learning has enabled robust sim-to-real locomotion for legged 
robots \cite{hwangbo2019learning,lee2020learning}, while massively parallel 
simulation has made the training of such policies practical 
\cite{rudin2022learning}. Periodic conditioning and reward composition 
have also been used to learn multiple bipedal gaits \cite{siekmann2021periodic}. 
We build on this general training framework and consider a coupled
loco-manipulation problem rather than command-tracking locomotion.
The policy must coordinate foot placement, balance, and upper-body 
forces with the rocking dynamics of a heavy external object.

\subsection{Humanoid Loco-Manipulation}

Humanoid systems have demonstrated constraint-based
loco-manipulation \cite{leziart2026cat}, whole-body
interaction with bulky objects \cite{zheng2025embracing}, and
transport of irregular loads using locomotion priors \cite{pan2026unwieldy}.
These approaches primarily lift, carry, embrace, place, or continuously 
push objects. In contrast, pivot walking repeatedly changes the object’s 
supporting ground corner while the robot continues to step. The policy must 
therefore coordinate locomotion with both contact forces and
the object’s alternating rocking motion.
\subsection{Pivoting and Nonprehensile Manipulation}

Pivoting is a classical nonprehensile manipulation primitive in which 
an object rotates about a supporting environmental contact rather than 
being fully supported by the robot \cite{aiyama1993pivoting}. 
Prior work has studied the mechanics and 
planning of pushing \cite{mason1986mechanics}, whole-body humanoid
pushing of large objects \cite{harada2003pushing,murooka2015pushing},
and pivot-based reorientation \cite{hou2018fast}.

Most closely related, Yoshida et al.~\cite{yoshida2010pivoting,yoshida2010planning}
demonstrated humanoid transport through planned quasi-static pivot sequences,
with object manipulation and robot stepping scheduled as separate phases.
Murooka et al.~\cite{murooka2014pivoting} subsequently added online 
object-state and force estimation to adapt planned humanoid pivoting
to motion error and unknown physical parameters. Zhang et al.~\cite{zhang2021graph} 
formulated pivoting-gait modes with graph model predictive control, 
though on a fixed industrial dual-arm manipulator rather than a walking
humanoid. Reinforcement learning has also produced \emph{in-hand} pivoting
\cite{antonova2017pivoting}, where a gripper supports the object and the
mobile base remains fixed.

In contrast, we propose a single whole-body humanoid policy that jointly 
generates stepping, balance, bimanual interaction, and alternating object 
rocking, without an explicit contact-mode schedule or a planned settling phase. 
The contribution is therefore the realization of pivoting as a
continuous learned humanoid gait rather than a new pivoting primitive.

\subsection{Reward Shaping and State-Indexed Motion References}
Complex locomotion and manipulation strategies can emerge from sparse
task objectives \cite{heess2017emergence, andrychowicz2020learning}. In
our setting, however, the outcome-only training routes tested in
Section~\ref{sec:exp_basic} converge to tilted dragging or unstable
pushing rather than alternating pivoting. We therefore use skill-level
reward shaping to make the desired object motion reachable during
exploration. Unlike potential-based shaping \cite{ng1999policy}, this
shaping intentionally changes the task optimum by specifying an
alternating box-rocking pattern, while leaving joint trajectories, foot
placement, contact timing, and interaction forces to be learned.

Phase-conditioned locomotion policies commonly use elapsed time or a
periodic clock \cite{siekmann2021periodic}. Conceptually related
state-based phase variables appear in virtual-constraint control, where
motion is indexed by robot configuration rather than time
\cite{westervelt2004experimental}. We instead index the box-yaw
reference by the object's realized forward progress. The reference
therefore pauses when the box stalls and reverses when it moves
backward, coupling the desired rocking phase to task progress rather
than elapsed time.
\section{Problem Formulation}
\label{sec:problem}

\subsection{Task Definition and Physical Setup}

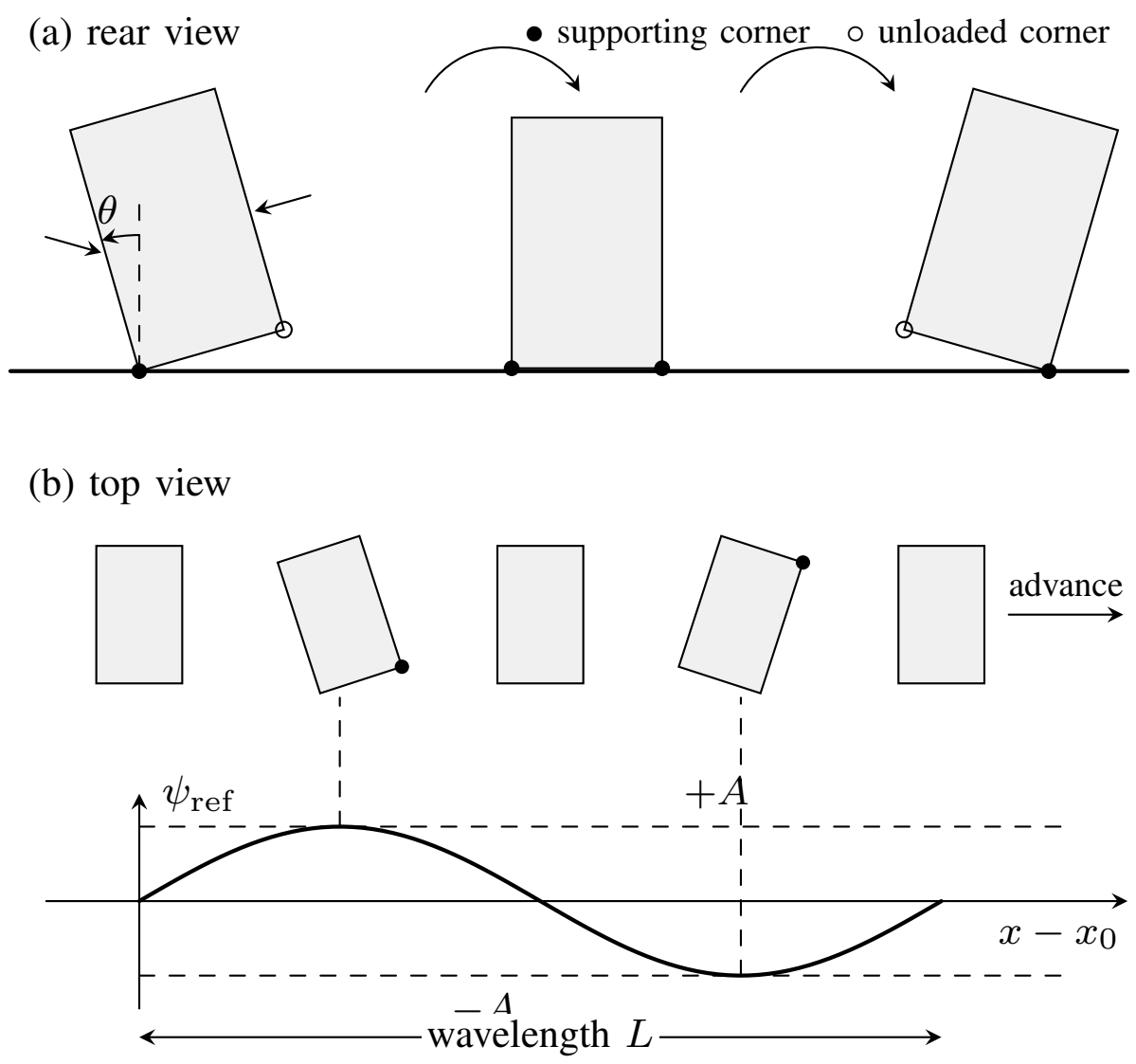
\begin{figure}[t]
  \centering
  \begin{tikzpicture}[>=stealth, font=\footnotesize, line cap=round]
    \begin{scope}
      \draw[thick] (-0.35,0) -- (7.45,0);
      \begin{scope}[shift={(0.55,0)}]
        \coordinate (p1bl) at (0,0);
        \begin{scope}[rotate around={16:(0,0)}]
          \draw[fill=black!6] (0,0) rectangle (1.05,1.75);
          \coordinate (p1br) at (1.05,0);
          \coordinate (p1l) at (0,0.85);
          \coordinate (p1r) at (1.05,0.85);
        \end{scope}
        \fill (p1bl) circle (0.055);
        \draw (p1br) circle (0.055);
        \draw[->] ($(p1l)+(-0.42,0.12)$) -- ($(p1l)+(-0.03,0.01)$);
        \draw[->] ($(p1r)+(0.42,0.12)$) -- ($(p1r)+(0.03,0.01)$);
        \draw[dashed] (0,0) -- (0,1.25);
        \draw[->] (0,0.95) arc[start angle=90, end angle=106,
          radius=0.95];
        \node at (-0.22,1.13) {$\theta$};
      \end{scope}
      \draw[->] (2.55,1.95) arc[start angle=150, end angle=30,
        radius=0.62];
      \begin{scope}[shift={(3.15,0)}]
        \draw[fill=black!6] (0,0.02) rectangle (1.05,1.77);
        \fill (0,0.02) circle (0.055);
        \fill (1.05,0.02) circle (0.055);
      \end{scope}
      \draw[->] (4.75,1.95) arc[start angle=150, end angle=30,
        radius=0.62];
      \begin{scope}[shift={(5.85,0)}]
        \coordinate (p3br) at (1.05,0);
        \begin{scope}[rotate around={-16:(1.05,0)}]
          \draw[fill=black!6] (0,0) rectangle (1.05,1.75);
          \coordinate (p3bl) at (0,0);
        \end{scope}
        \fill (p3br) circle (0.055);
        \draw (p3bl) circle (0.055);
      \end{scope}
      \node[anchor=west] at (-0.35,2.35) {(a) rear view};
      \node[anchor=east, align=right] at (7.45,2.35)
        {\scriptsize $\bullet$ supporting corner\quad
         $\circ$ unloaded corner};
    \end{scope}
    \begin{scope}[shift={(0,-1.15)}]
      \node[anchor=west] at (-0.35,0.35) {(b) top view};
      \foreach \cx/\ang/\marks in
        {0.55/0/none, 1.95/18/left, 3.35/0/none, 4.75/-18/right,
         6.15/0/none}{
        \begin{scope}[shift={(\cx,-0.55)}, rotate=\ang]
          \draw[fill=black!6] (-0.30,-0.48) rectangle (0.30,0.48);
        \end{scope}
      }

      \fill ($(1.95,-0.55)+({0.30*cos(18)+0.48*sin(18)},
        {0.30*sin(18)-0.48*cos(18)})$) circle (0.05);
      \fill ($(4.75,-0.55)+({0.30*cos(-18)-0.48*sin(-18)},
        {0.30*sin(-18)+0.48*cos(-18)})$) circle (0.05);

      \draw[->] (6.62,-0.55) -- (7.42,-0.55)
        node[midway, above] {\scriptsize advance};

      \begin{scope}[shift={(0,-2.55)}]
        \draw[->] (-0.1,0) -- (7.45,0) node[below left=0pt and -2pt]
          {$x-x_0$};
        \draw[->] (0.55,-0.75) -- (0.55,0.75)
          node[right=1pt] {$\psi_{\mathrm{ref}}$};
        \draw[thick] plot[domain=0:5.6, samples=80]
          ({0.55+\x}, {0.52*sin(360*\x/5.6)});
        \draw[dashed] (0.55,0.52) -- (7.05,0.52)
          node[pos=0.62, above] {$+A$};
        \draw[dashed] (0.55,-0.52) -- (7.05,-0.52)
          node[pos=0.37, below] {$-A$};
        \draw[dashed] (1.95,0.52) -- (1.95,1.42);
        \draw[dashed] (4.75,-0.52) -- (4.75,1.42);
        \draw[<->] (0.55,-0.95) -- (6.15,-0.95)
          node[midway, fill=white, inner sep=1pt]
          {wavelength $L$};
      \end{scope}
    \end{scope}
  \end{tikzpicture}
  \caption{Pivot-walking schematic (not to scale). (a) Rear view: the
  box rocks between its two bottom corners, so one corner supports the
  box while the opposite corner is unloaded, and the hands press on the
  side faces to steer the rotation. (b) Top view: box yaw alternates
  with amplitude $A$ as the box advances, transferring support once per
  half wavelength. The yaw reference $\psi_{\mathrm{ref}}$ used for
  training (Section~\ref{sec:method_path}) is indexed by box progress
  $x$, so it halts whenever the box halts.}
  \label{fig:concept}
\end{figure}

We consider whole-body transport of a free-standing box by a 29-DoF
Unitree G1 EDU humanoid (legs, waist, arms, wrists). The goal is to
learn a policy that observes the robot state and, during the main
training stage, the box state, and outputs whole-body joint targets
that move the box forward. The transport axis is fixed at episode
initialization, and the box yaw should stay close to its initial
heading. The robot receives no walking velocity command, no footstep
sequence, and no contact schedule. The desired strategy keeps both
hands on the two side faces of the box and repeatedly rocks it
between its two bottom corners while alternating its yaw, which
produces forward motion (Fig.~\ref{fig:concept}).

Simulation uses mjlab, built on MuJoCo Warp for GPU-parallel
training. The physical object is a modified G1 shipping case. It
measures $0.475\times0.35\times\SI{0.82}{m}$ (width $\times$ depth
$\times$ height) and weighs $\SI{14.9}{kg}$. Simulation models it as
a rigid free-standing cuboid (hardware preparation in
Section~\ref{sec:exp_real}). The robot starts in a crouched carry
pose with a nominal base height of $\SI{0.765}{m}$, and each palm
touches one side face of the box near mid-height. By pressing the two
side faces with different forces, the hands can rotate the box in yaw
while the ground carries most of its weight. Hand contact is
encouraged by the reward and is not enforced as a constraint.

If the box were lifted with the load shared equally, each arm would
carry $\SI{7.45}{kg}$. This is more than twice the $\SI{3}{kg}$
per-arm payload of the G1 EDU \cite{unitree2024g1}. Dragging, in
turn, requires greater force as friction increases and can abrade the
object and the floor. We therefore study ground-supported pivot
transport for this robot and object. Box mass and box--ground
friction vary across training and evaluation, and the exact values
are given in Sections~\ref{sec:method} and~\ref{sec:experiments}.

\subsection{Policy Observations and Actions}
At each control step the policy maps an observation to an action,
$a_t = \pi_\theta(o_t)$. The action sets joint-position targets for
all 29 actuated joints as offsets from a nominal pose. The policy
runs at $\SI{50}{Hz}$ and a low-level PD controller tracks the
targets. The box-observing training actor receives robot
proprioception (joint positions and velocities, base angular
velocity, and projected gravity), the previous action, and the box
state relative to the robot base. The box state contains the relative
position, the sine and cosine of the yaw deviation from the initial
heading, and gravity projected into the box frame. Skill-reward
training adds a three-dimensional reference-phase observation
(Section~\ref{sec:method_shaped}). No walking velocity command is
given. Uniform noise is added to each actor observation term in raw
units, for example $\pm0.05$ on the box terms. The critic receives
the noise-free state, including variables that the deployment actor
cannot observe. We study two actor settings. The box-observing policy
above is used for gait learning. The object-blind deployment actor
removes all box observations and instead receives a periodic phase
signal and the robot yaw from gait onset
(Section~\ref{sec:method_blind}).

\subsection{Episode and Termination}
Each simulation episode lasts $\SI{20}{s}$ and terminates early if
the box tilt exceeds $\ang{35}$, the robot falls (base orientation
more than $\ang{70}$ from upright), or a foot contacts the box.
Foot--box contact is treated as failure because the task is defined
as bimanual pivot transport rather than leg-assisted pushing. Episode
completion \rev{measures consistency across initial box orientations
($\pm\ang{5}$, one box).} Transport distance and strategy-level metrics are evaluated
separately, because forward progress alone cannot distinguish
alternating-corner pivoting from tilted dragging. The corresponding
gait-validity metrics are defined in
Section~\ref{sec:method_metrics}.
\section{Method}
\label{sec:method}

\subsection{Overview}
\label{sec:method_basic}
\begin{figure*}[t]
  \centering
  \begin{tikzpicture}[
    stage/.style={draw, align=center, font=\footnotesize,
      inner sep=4pt, minimum height=0.5in, text width=0.16\textwidth},
    >=stealth, node distance=0.26in and 0.24in]
    \node[stage] (stance) {\textbf{Stance pretraining}\\
      balance while holding\\ a tilted box};
    \node[stage, right=of stance] (clockg) {\textbf{Clock-gait
      discovery}\\ skill rewards,\\ $\SI{4}{s}$ temporal clock};
    \node[stage, right=of clockg] (spath) {\textbf{Spatial-path
      conversion}\\ $\psi_{\mathrm{ref}}(x)$ paced by\\ box progress};
    \node[stage, right=of spath] (mass) {\textbf{Mass ladder}\\
      $4\to8\to\SI{14.9}{kg}$\\ box-observing $\pi_{\mathrm{path}}$};
    \node[stage, below=of spath] (blind) {\textbf{Object-blind
      transfer}\\ input-layer modification\\ + continued RL,
      $\pi_{\mathrm{clock}}$};
    \node[stage, right=of blind] (hw) {\textbf{Hardware
      continuation}\\ hardware-centered\\ randomization,
      $\pi_{\mathrm{hw}}$};
    \node[stage, dashed, below=of stance] (outcome)
      {\textbf{Outcome-only control}\\ progress + hand contact\\
      $\Downarrow$ tilted dragging};
    \draw[->] (stance) -- (clockg);
    \draw[->] (clockg) -- (spath);
    \draw[->] (spath) -- (mass);
    \draw[->] (mass.south) -- ++(0,-0.12in) -| (blind.north);
    \draw[->] (blind) -- (hw);
  \end{tikzpicture}
  \caption{Training pipeline. The main route (solid) pretrains a
  stance, discovers the alternating gait under a temporal clock,
  converts the reference to a progress-paced spatial path, and adapts
  the gait up the mass ladder. The resulting box-observing policy
  $\pi_{\mathrm{path}}$ is transferred to an object-blind actor
  $\pi_{\mathrm{clock}}$, and a hardware-centered continuation
  $\pi_{\mathrm{hw}}$ runs in the physical experiments
  (Section~\ref{sec:exp_real}). The outcome-only control (dashed) is
  trained independently for comparison, not used as an initialization.}
  \label{fig:lineage}
\end{figure*}
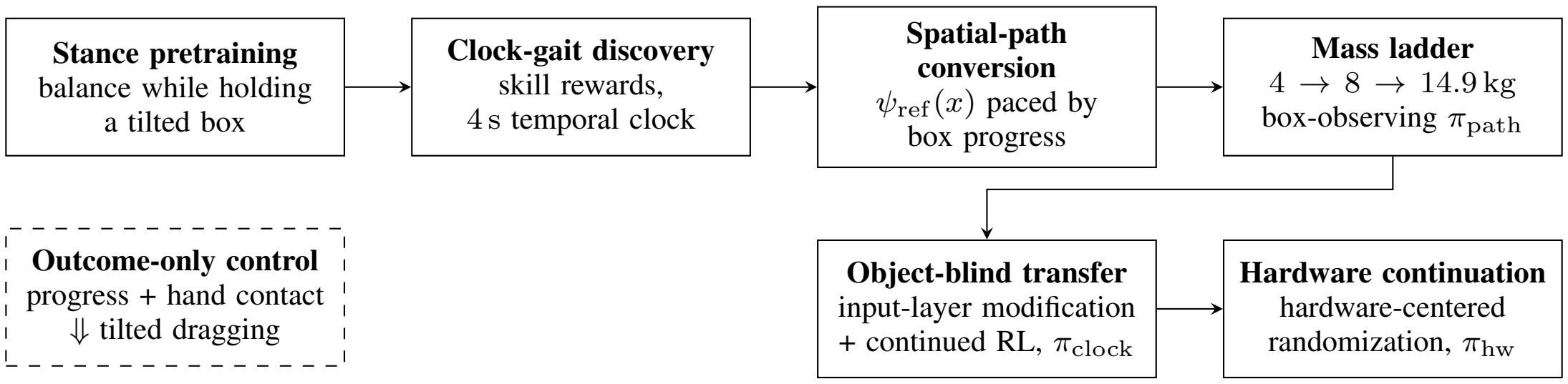

Our goal is a single whole-body policy that advances a ground-supported
box through repeated alternating pivot motions. The policy controls all
29 robot joints and must coordinate box rocking, bimanual side contact,
balance, and stepping without an explicit contact-mode schedule.
Fig.~\ref{fig:lineage} summarizes the training pipeline.

The method has four stages. First, the robot learns to balance while
keeping side contact with a tilted, ground-supported box. Second, a
temporal yaw reference and box-level skill rewards let it discover an
alternating pivot gait at low box mass. Third, the temporal reference
is replaced by a spatial reference indexed by realized box progress,
and the gait is adapted through a mass curriculum. Finally, box-state
observations are removed and the policy is trained further with an
internal clock for hardware deployment.

\subsection{Pivot-Gait Reward}
\label{sec:method_shaped}
The basic transport objective rewards achieved forward displacement
while penalizing persistent box-heading deviation:
\begin{equation}
  r_{\text{fwd}} = \operatorname{clip}\!\Big(\tfrac{x^{\text{box}} -
  x_0}{x^\ast - x_0}, 0, 1\Big)\cdot
  \exp\!\big(-(\psi^{\text{box}}/\sigma_\psi)^2\big),
  \label{eq:outcome}
\end{equation}
where $x^{\text{box}}$ is the box position along the fixed transport
axis, $x_0$ its initial value, $x^\ast$ the nominal progress
target, $\psi^{\text{box}}$ the box yaw deviation, and $\sigma_\psi =
\ang{30}$. This state reward is paid at every step, so earlier
progress yields a larger return. The yaw factor discourages
persistent heading drift while permitting the alternating yaw
excursions needed for pivot walking. A hand--box contact reward and
regularization terms for posture, action rate, and upright base
orientation complete the basic objective, and early termination
incurs a penalty of $-200$. In the outcome-only routes of
Section~\ref{sec:exp_basic}, this objective converges to tilted
dragging or unstable pushing instead of repeated pivoting. We
therefore add box-level skill rewards.

For gait discovery, a Gaussian reward tracks a temporal yaw
reference, a simple periodic target for exploration:
\begin{equation}
  r_{\text{osc}} = \exp\!\Big(\!-\big(\tfrac{\psi^{\text{box}} -
  \psi_{\mathrm{ref}}}{\sigma_{\mathrm{osc}}}\big)^{2}\Big),\qquad
  \psi_{\mathrm{ref}}(t) = A\sin\!\big(\tfrac{2\pi t}{T}\big),
  \label{eq:osc}
\end{equation}
with period $T=\SI{4}{s}$ and width $\sigma_{\mathrm{osc}} =
\max(0.6A, \ang{6})$. The amplitude $A$ grows from $\ang{5}$ to
$\ang{25}$ over roughly the first 2000 training iterations, and the
actor observes the sine and cosine of the reference phase together
with the current amplitude factor. A binary tilt reward encourages
configurations associated with bottom-corner support:
\begin{equation}
  r_{\text{tilt}} =
  \begin{cases}
    1, & \ang{12} < \theta < \ang{30},\\
    0, & \text{otherwise},
  \end{cases}
  \label{eq:tiltband}
\end{equation}
where $\theta$ is the angle between the box up-axis and the world
vertical. The lower bound discourages flat-ground rotation, and the
upper bound leaves margin before the $\ang{35}$ termination threshold.
This term does not prescribe which corner is loaded. Alternating
support arises from its combination with the yaw reference. An
additional reward keeps each hand near its assigned side face, and a
small torso--box contact penalty prevents the robot from leaning on
the object. All terms are combined additively, because a
multiplicative combination gives zero reward in the initial upright
state and removes the incentive to start moving. The reference
specifies only the desired object motion. Joint trajectories, foot
placements, contact timing, and hand forces are all learned.

\subsection{Progress-Paced Spatial Reference}
\label{sec:method_path}
A temporal reference continues to advance even when a heavy box
stalls, so the clock demands yaw reversals that the realized box
motion has not reached. We therefore index the reference by box
progress instead of time:
\begin{equation}
  \psi_{\mathrm{ref}}(x) = A\sin\!\Big(\tfrac{2\pi\,(x -
  x_0)}{L}\Big),\qquad A=\ang{18},\; L=\SI{0.20}{m}.
  \label{eq:path}
\end{equation}
The desired yaw changes sign every $L/2 = \SI{0.10}{m}$ of box
progress. When the box stops, the reference phase also stops, and
backward motion reverses the phase.

A stationary box makes the spatial reference constant, so yaw
tracking alone can be maximized without transport. We therefore
multiply the tracking term by capped positive box velocity:
\begin{equation}
  r_{\text{path}} = \exp\!\Big(\!-\big(\tfrac{\psi^{\text{box}} -
  \psi_{\mathrm{ref}}(x)}{\ang{6}}\big)^{2}\Big)\cdot
  \operatorname{clip}\big(\dot{x}^{\text{box}}, 0, v_{\max}\big),
  \label{eq:pathprog}
\end{equation}
where $v_{\max} = \SI{0.15}{m/s}$ and $\dot{x}^{\text{box}}$ is the
box velocity along the transport axis. The cap saturates the reward
at higher speeds and is not a hard speed limit. The velocity factor
is left unnormalized, and its scale is absorbed into the reward
weight. This term replaces the temporal tracking term, while the
other rewards remain active.

For heavy-box training we add an overspeed penalty on a $\SI{2}{s}$
moving average of box speed and a quadratic penalty on base-height
deviation. The average tolerates the short velocity peaks of the
rocking gait, and the height term maintains the nominal crouch.

The actor observes the sine and cosine of $2\pi(x - x_0)/L$, the same
two-dimensional phase embedding as the clock, so training resumes
directly across the switch. We denote the resulting box-observing
policy $\pi_{\mathrm{path}}$.

\subsection{Training Curriculum}
\label{sec:method_curriculum}
Training uses PPO with 4096 parallel environments. The actor and critic
are multilayer perceptrons with hidden dimensions $(512,256,128)$ and
ELU activations. Each iteration collects 24 policy steps per
environment and performs five optimization epochs, with $\gamma=0.99$,
GAE $\lambda=0.95$, and a KL-adaptive learning rate. The policy is
trained through four stages, each resuming the previous one.
\begin{enumerate}
  \item \textbf{Stance pretraining.} At $\SI{4}{kg}$, the robot learns
  to stay balanced while keeping side contact with a tilted,
  ground-supported box, with progress rewards disabled and box tilt
  and yaw randomized at reset.
  \item \textbf{Pivot-gait discovery.} The temporal reference and
  skill rewards are activated at $\SI{4}{kg}$ together with the
  amplitude curriculum. Alternating yaw swings appear within
  approximately 2000 iterations.
  \item \textbf{Spatial-reference conversion.} The temporal phase is
  replaced by the progress-paced reference of Eq.~\eqref{eq:path} and
  training continues.
  \item \textbf{Mass adaptation.} The gait is adapted from 4 to 8 and
  then $\SI{14.9}{kg}$, with a separate continuation to $\SI{24}{kg}$.
  The rewards and reference stay fixed apart from the heavy-box
  regularization of Section~\ref{sec:method_path}.
\end{enumerate}
We use this staged initialization because direct training at the
heavier masses did not produce a stable gait in our runs.

\subsection{Object-Blind Deployment Transfer}
\label{sec:method_blind}
The policy $\pi_{\mathrm{path}}$ requires the box pose and the box
progress that indexes its phase. Neither is available on our hardware,
which has no motion capture or box-mounted sensing. We therefore
transfer the gait to an actor that uses onboard measurements only, by
modifying the actor input layer and continuing RL training.

The box-state columns of the actor's first layer are removed and all
remaining weights are kept. The critic retains the box state as
privileged information during continued training. We also append a
zero-initialized input carrying the sine and cosine of the robot yaw
from gait onset, which reduces slow heading drift.

Because box progress is no longer observed, the spatial phase is
replaced by a periodic clock, and the spatial-path term is replaced by
a capped forward-speed reward. The box velocity in this reward is
computed by the simulator, so removing box sensing from the actor does
not restrict training. The learned gait has a full yaw cycle of about
$\SI{1.0}{s}$ at $\SI{14.9}{kg}$, and for hardware-oriented training
we choose a slower $\SI{1.4}{s}$ clock period to temper the motion.
The resulting actor $\pi_{\mathrm{clock}}$ observes joint positions
and velocities (29 each), base angular velocity and projected gravity
(3 each), the previous action (29), the three-dimensional clock
signal, and the robot-yaw pair, 98 scalars in total, all available
onboard.

Continued training randomizes box mass, box--ground and hand friction,
box center of mass, initial box placement, PD gains, and hand-contact
geometry. A slow-start factor ramps the rocking amplitude and the
speed reward over the first $\SI{6}{s}$ of each episode. A final
continuation adds symmetry and heading-drift regularization and
re-centers the friction randomization on the measured floor. This
policy, denoted $\pi_{\mathrm{hw}}$, is used in the physical
experiments (Section~\ref{sec:exp_real}).
\section{Experiments}
\label{sec:experiments}

\subsection{Evaluation Protocol and Metrics}
\label{sec:method_metrics}
Unless stated otherwise, each simulation result evaluates the
deterministic actions of one trained checkpoint for $\SI{20}{s}$ in 256
parallel environments (128 for the $\SI{14.9}{kg}$ path policy). The
episodes randomize only the initial box orientation, with yaw, roll,
and pitch sampled within $\pm\ang{5}$, not the training seed or the
physical parameters. Reported forward travel is accumulated positive
travel and can exceed net displacement, and ``terminations'' counts
robot falls, box falls, and foot--box events. Training rewards
progress toward a nominal $\SI{2}{m}$ target (Eq.~\ref{eq:outcome}).
Evaluation instead runs for $\SI{20}{s}$ without a stop command, so
travel beyond $\SI{2}{m}$ shows gait continuation rather than
controlled delivery.

Because scalar task metrics cannot separate pivot walking from tilted
dragging, we also count amplitude-segmented yaw swings (excursions over
$\ang{10}$ between sign changes) and report the lowest-corner motion
proxy, the integrated planar speed of the lowest box corner divided by
forward travel. The proxy is near zero for rotation about a stationary
corner and near one for pure translation; it does not identify the
active contact. For selected checkpoints we therefore replay
deterministic trajectories in MuJoCo and measure per-corner
box--ground contact forces, from which we report left--right support
alternations, the fraction of time one corner carries over $60\%$ of
the normal force, and the loaded-corner slide per forward travel.
Swings, support alternations, and hardware yaw reversals
(Section~\ref{sec:exp_real}) use different detectors and are not
interchangeable counts.

\subsection{Does Outcome-Only Training Discover Pivot Walking?}
\label{sec:exp_basic}
None of the six outcome-only routes (one at $\SI{8}{kg}$, one at
$\SI{14.9}{kg}$, four at $\SI{24}{kg}$, one training seed each)
produces repeated alternating yaw swings. Five routes converge to
stable constant-tilt dragging, holding a tilt of
$\ang{25}$--$\ang{31}$ and advancing only $0.18$--$\SI{0.33}{m}$ per
episode. The sixth is an unstable deep-tilt push that repeatedly
topples the box. These routes show that the gait did not emerge under
the tested settings, not that it cannot emerge, and they motivate the
explicit pivoting objective.
\rev{To test this within our pipeline, we train two branches from the
same $\SI{4}{kg}$ stance-pretrained policy for 6000 updates. The only
difference is the box-level skill rewards. With them, the robot learns
repeated yaw swings (11.4 per episode), although it still often fails
at this early stage. Without them, it learns no swings and often ends
with a foot touching the box. In this single-seed test, only the
branch with skill rewards learned the gait.}

\subsection{Does the Learned Policy Actually Pivot the Box?}
\label{sec:exp_ladder}

\begin{figure*}[t]
  \centering
  \includegraphics[width=0.98\textwidth]{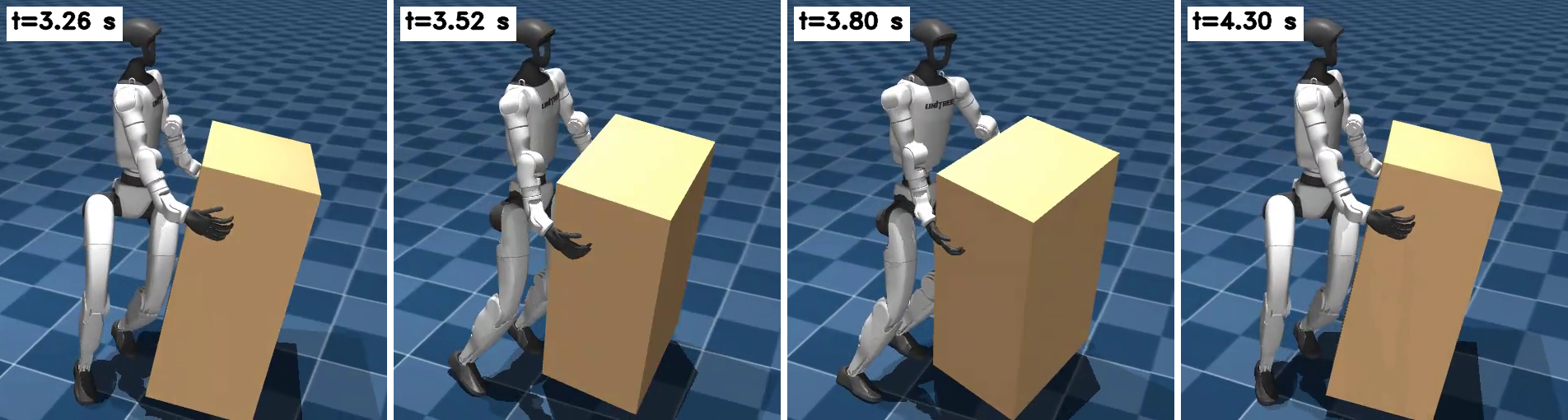}
  \caption{$\pi_{\mathrm{post}}$, a posture-refined sibling of
  $\pi_{\mathrm{path}}$, at $\SI{14.9}{kg}$
  (a separate checkpoint from Table~\ref{tab:ladder}). Four frames span
  one full alternating yaw cycle ($+\ang{17}$ peak, transition between
  lean directions near zero yaw, $-\ang{17}$ peak, return) while the box
  stays leaned onto its near edge; evaluation records
  $\SI{3.68}{m}$/$\SI{20}{s}$ and zero terminations over 256
  environments.}
  \label{fig:simulation_sequence}
\end{figure*}

\begin{table}[t]
  \centering
  \caption{Path-conditioned pivot gait across the mass curriculum
  ($\mu_g{=}2$, $\SI{20}{s}$). Values are means across 256 evaluated
  initial conditions (128 at $\SI{14.9}{kg}$); $T_{\mathrm{cycle}}=2(\SI{20}{s})/N_{\mathrm{swing}}$;
  LC motion is the normalized lowest-corner motion proxy. All rows
  complete the episode with zero terminations.}
  \label{tab:ladder}
  \begin{tabular}{rrrrr}
    \toprule
    Mass & Travel & Swings & $T_{\mathrm{cycle}}$ & LC motion \\
    {[kg]} & [m] & [/env] & [s] & \\
    \midrule
    4 & 5.88 & 58.8 & 0.68 & 0.57 \\
    8 & 5.45 & 54.5 & 0.73 & 0.36 \\
    \textbf{14.9} & \textbf{4.12} & \textbf{41.2} & \textbf{0.97} &
      \textbf{0.059} \\
    \bottomrule
  \end{tabular}
\end{table}

\textbf{Main result.} Table~\ref{tab:ladder} evaluates the
path-conditioned policy $\pi_{\mathrm{path}}$ across the mass
curriculum. At the physical mass of $\SI{14.9}{kg}$ it completes all
128 episodes without termination and accumulates $\SI{4.12}{m}$ of
forward travel in $\SI{20}{s}$ ($\SI{0.21}{m/s}$), with 41.2 yaw
swings per episode (mean amplitude $\ang{32.6}$). Nearly all of this
travel ($98.7\%$) occurs during the swings. The full yaw-cycle period
lengthens from $\SI{0.68}{s}$ at $\SI{4}{kg}$ to $\SI{0.97}{s}$ at
$\SI{14.9}{kg}$, although the reference carries no time schedule.
With a fixed spatial wavelength, the slowing follows the lower
transport speed. The lowest-corner proxy falls to $0.059$, consistent
with a nearly stationary supporting corner.

\rev{In simulation, the feet stay low: the soles are within
$\SI{15}{mm}$ of the floor in $99.0\%$/$98.4\%$ of left/right samples
and rise above this height only $1.4$/$2.9$ times per episode, indicating
shuffling rather than distinct steps.}

\textbf{Contact-force validation.} Replaying $\pi_{\mathrm{post}}$, a
posture-refined sibling of $\pi_{\mathrm{path}}$
(Fig.~\ref{fig:simulation_sequence}), in deterministic MuJoCo, we
measure box--ground contact forces directly. Over $\SI{20}{s}$ the
dominant support alternates 34 times between the left and right bottom
corners, a single corner is dominant during $99.5\%$ of the rollout,
and the loaded-corner slide is $0.09$ of forward travel
(Fig.~\ref{fig:contact_timeline}). The learned motion is therefore
alternating-corner pivoting rather than tilted translation, for the
replayed checkpoint and its contact model. \rev{Replaying
$\pi_{\mathrm{path}}$ itself gives 32 alternations, $95.7\%$
single-corner dominance, and slide $0.123$, although it travels less in
this replay ($\SI{2.32}{m}$) than in training simulation.}

\textbf{Seed replication.} Re-running the pipeline with two additional
training seeds \rev{also} reproduces the \rev{gait}. All three seeds
discover the gait at $\SI{4}{kg}$, complete the curriculum with zero
terminations, transport $4.12$--$\SI{4.32}{m}$ with $41$--$43$ swings,
and slow with mass in the same way. Only the proxy is seed-dependent
($0.059$, $0.29$, $0.054$); the reproducible outcome across these
seeds is the alternating gait, not consistently low corner motion.

\subsection{How Do Temporal and Progress-Paced Routes Compare?}
\label{sec:exp_ablation}

\begin{table}[t]
  \centering
  \caption{Complete reference-conditioning routes at $\SI{14.9}{kg}$
  ($\SI{20}{s}$). 
  ($^{*}$computed over rollouts with repeated box-fall resets).}
  \label{tab:ablation}
  \footnotesize
  \setlength{\tabcolsep}{2.8pt}
  \begin{tabular}{llrrrr}
    \toprule
    Reference & Init & Travel & Swings & Term. & LC \\
     & & [m] & [/env] & & \\
    \midrule
    clock $\SI{4}{s}$ (hand-chosen) & mass ladder & 0.84 & 10.4 & 0 & 0.23 \\
    clock $\SI{1.1}{s}$ & $\SI{4}{s}$-gait policy & 1.51 & 36.0 & 0 & 0.18 \\
    clock $\SI{1.1}{s}$ & scratch + ladder & 4.83 & 35.9 & 963 &
      0.08$^{*}$ \\
    \textbf{path (progress-paced)} & mass ladder & \textbf{4.12} &
      \textbf{41.2} & \textbf{0} & \textbf{0.059} \\
    \bottomrule
  \end{tabular}
\end{table}

Table~\ref{tab:ablation} compares four complete training routes at
$\SI{14.9}{kg}$. The hand-chosen $\SI{4}{s}$
clock is stable yet slow ($\SI{0.84}{m}$). Retuned to
$\SI{1.1}{s}$, the clock route reaches $\SI{1.51}{m}$, and a from-scratch
$\SI{1.1}{s}$ clock reaches $\SI{4.83}{m}$ while toppling the box 963
times across 256 environments. The progress-paced route transports
$\SI{4.12}{m}$ with zero terminations. The results are consistent
with progress coupling
aiding synchronization for the heavier box. \rev{To control for
initialization, we continued $\pi_{\mathrm{path}}$ for 1200 updates under
its path phase or a $\SI{0.97}{s}$ clock; both branches lost the gait,
with no yaw swings, so this comparison does not separate the two
references.}

\subsection{How Does Performance Change with Box Mass?}
\label{sec:exp_heavy}
At $\SI{24}{kg}$, under the same side-contact configuration and
$\mu_g{=}2$, the four outcome-only routes of
Section~\ref{sec:exp_basic} either drag ($0.18$--$\SI{0.33}{m}$) or
topple the box ($\SI{0.42}{m}$ with 1003 box falls). Adapting the
pivot policy with the skill objective instead transports
$0.76$--$\SI{0.81}{m}$ with $10$--$13$ swings and zero terminations
across two seeds, versus at most $\SI{0.33}{m}$ for the failure-free
draggers. Reducing the skill-reward weights below half lets the gait
decay toward dragging. A probe task that rewards only rotating the
box to a $\ang{30}$ yaw succeeds at $\SI{24}{kg}$ ($0.99$ tracking)
and fails at $\SI{28}{kg}$, bracketing an empirical boundary of the
tested training procedure and contact configuration, not a physical
limit.

\subsection{Can the Gait Be Transferred without Box-State Sensing?}
\label{sec:exp_sim2sim}

\begin{figure}[t]
  \centering
  \includegraphics[width=0.99\columnwidth]{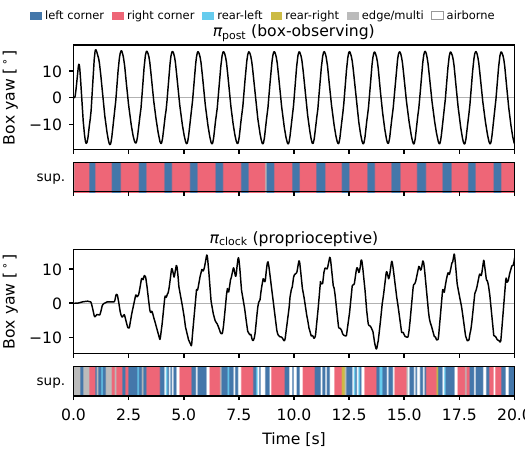}
  \caption{Box yaw and dominant support corner from measured
  box--ground contact forces in deterministic replay}
  \label{fig:contact_timeline}
\end{figure}

\begin{figure*}[t]
  \centering
  \includegraphics[width=0.98\textwidth]{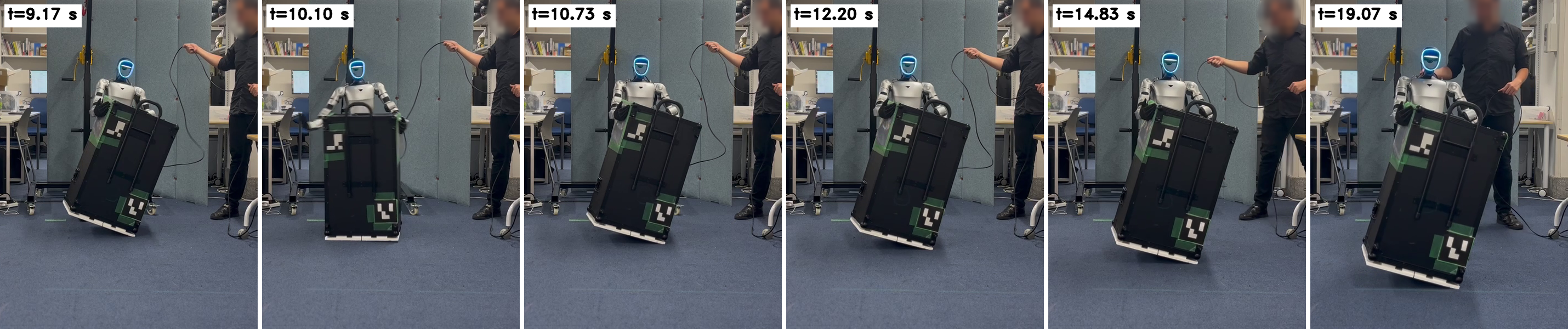}
  \caption{Transfer of real robot experiment of the alternating gait by
  $\pi_{\mathrm{hw}}$ (trial 3, fixed frontal camera; timestamps
  from the marker track): a lean peak ($t{=}\SI{9.17}{s}$), the return
  through near-upright at corner transfer ($\SI{10.10}{s}$), the opposite
  lean ($\SI{10.73}{s}$), and continued alternation as the box advances
  toward the camera; the active contact corner is not directly measured,
  and the uncut recordings include the safety-triggered trial ending.}
  \label{fig:hardware_sequence}
\end{figure*}

In the training simulator, the clock-conditioned actor
$\pi_{\mathrm{clock}}$ (Section~\ref{sec:method_blind}) travels
$\SI{4.67}{m}$ in the $\SI{20}{s}$ evaluation with zero terminations.
\rev{In the same simulator, the deployed policy $\pi_{\mathrm{hw}}$ travels
$\SI{4.44}{m}$ with about 26 swings, and the box falls in 4 of 512
episodes. Its corner proxy is high ($0.66$), so we cannot confirm
alternating support for this policy. Its feet stay within $\SI{15}{mm}$
of the floor in $86\%$/$91\%$ of left/right samples. Foot motion was not
measured on hardware.}
\rev{The exported $\pi_{\mathrm{path}}$ and $\pi_{\mathrm{clock}}$ policies} also run unchanged in an independent
MuJoCo deployment stack with the C++ controller used on the robot.
With simulated box pose, $\pi_{\mathrm{path}}$ travels $\SI{10.2}{m}$
in $\SI{55}{s}$ without falls, confirming backend transfer.
$\pi_{\mathrm{clock}}$, which receives no box pose or progress
signal, travels $\SI{9.9}{m}$ in its $\SI{73.8}{s}$ active segment
with zero falls, leaving the box upright at the commanded stop. It
drifts $\SI{1.70}{m}$ laterally\rev{, consistent with uncorrected phase
or heading error}.

Contact replay shows that the alternating support survives the loss of
sensing. $\pi_{\mathrm{clock}}$ produces 30 left--right support
alternations in $\SI{20}{s}$, close to the ${\sim}29$ its
$\SI{1.4}{s}$ clock predicts, with one corner dominant $78\%$ of the
time and a loaded-corner slide of $0.24$ of forward travel, against
$0.09$ for the box-observing $\pi_{\mathrm{post}}$
(per-policy values, not a head-to-head comparison;
Fig.~\ref{fig:contact_timeline}). \rev{For $\pi_{\mathrm{clock}}$,} the
gait survives the loss of sensing, with more sliding and drift.

\begin{figure}[t]
  \centering
  \includegraphics[width=1.0\columnwidth]{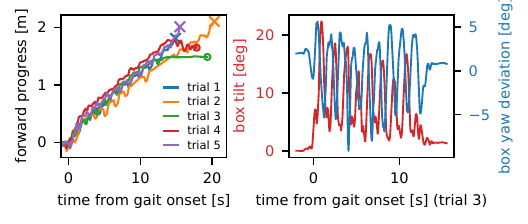}
  \caption{Marker-tracked real robot trials. Left: estimated forward
  progress from gait onset for all five trials (crosses mark trials
  ending with box tilt above $\ang{30}$, circles mark lower final
  tilt). Right: estimated box tilt and yaw for trial 3, showing
  repeated lean and yaw reversals during forward motion. Marker
  estimates are approximate and do not identify the active
  ground-contact corner; the relative progress shown is not the
  tape-measured advance of Table~\ref{tab:hw_trials}.}
  \label{fig:hw_trials}
\end{figure}

\subsection{Does the Gait Transfer to Hardware?}
\label{sec:exp_real}
The hardware experiment evaluates transfer of the alternating gait,
not delivery or set-down. We ran five filmed trials on the physical
G1 with the $\SI{14.9}{kg}$ case of Section~\ref{sec:problem},
prepared with casters removed, handle covers flattened, and rubber
sheets at the hand-contact regions, and report all five. The
deployed actor
$\pi_{\mathrm{hw}}$ is the hardware continuation of
Section~\ref{sec:method_blind} and runs on the onboard inputs only,
with the same observations and $\SI{1.4}{s}$ clock as
$\pi_{\mathrm{clock}}$. \rev{Section~\ref{sec:exp_sim2sim} reports $\pi_{\mathrm{hw}}$ in
simulation; this campaign is its hardware evaluation.} Box motion is estimated offline
from two front-face markers and is used only for gait timing and
relative progress. The one physically measured distance is the
tape-measured advance in Table~\ref{tab:hw_trials}.

\begin{table}[t]
  \centering
  \caption{Repeated hardware trials ($n{=}5$).}
  \label{tab:hw_trials}
  \begin{tabular}{lr}
    \toprule
    Alternating gait reproduced & $5/5$ \\
    Observed gait duration [s] & $16.2 \pm 3.0$ \\
    Active-gait marker speed [m/s] & $0.11 \pm 0.01$ \\
    Detected yaw reversals [/trial] & $15.6 \pm 3.0$ \\
    Start--final-rest advance [m] & $1.93 \pm 0.41$ \\
    \midrule
    Base-tilt safety trigger ended trial & $5/5$ \\
    \rev{Robot fell after Passive transition} & \rev{$5/5$} \\
    Box toppled during triggered ending & $2/5$ \\
    \bottomrule
  \end{tabular}
\end{table}

The alternating gait is reproduced in all five trials
(Table~\ref{tab:hw_trials}, Figs.~\ref{fig:hardware_oblique},
\ref{fig:hardware_sequence}, and~\ref{fig:hw_trials}). The box advances
through repeated lean and yaw reversals, with tilt swings averaging
$\ang{16}$ (peaks $22$--$\ang{34}$) at a cadence commensurate with the
$\SI{1.4}{s}$ clock; the motion is qualitatively consistent with the
alternating-corner support measured in simulation, though the contact
corner is not instrumented. All five trials end $12$--$\SI{22}{s}$
after gait onset, when base tilt exceeds $\SI{1.0}{rad}$ and the
controller drops to Passive\rev{, after which the robot collapses}; no box fall occurred during the preceding
gait, and the box toppled in two trials during this triggered ending.
The endings are systematic (a same-signed final yaw excursion in all
five tracks and the same backward crouching collapse in the side
views). \rev{The yaw drift is consistent with accumulated phase or heading
error in the object-blind controller. However, we did not log joint and
IMU data, so we could not identify what makes the base tilt grow.} The campaign
demonstrates repeatable short-horizon transfer of the gait; sustained
delivery, direct measurement of support-corner alternation, and a
controlled stop remain undemonstrated.
\section{Conclusion}
\label{sec:conclusion}

In summary, we presented a learned whole-body pivot-walking gait that
jointly coordinates humanoid stepping, balance, and bimanual
manipulation of a ground-supported box. A spatial reference paced by the
realized box progress couples the alternating box motion to actual
transport \rev{during training. The deployed policy instead follows a
periodic clock}.

In the six outcome-only routes we tested, a transport reward alone does
not produce the gait. With box-level pivoting rewards, the policy
transports a $\SI{14.9}{kg}$ box by an average of $\SI{4.12}{m}$ in
$\SI{20}{s}$ over 128 \rev{initial box orientations ($\pm\ang{5}$, one box)}
with no observed terminations, and
contact-force replay of a posture-refined continuation verifies repeated
alternation of the dominant support corner. On hardware, an object-blind
continuation using only onboard observations reproduces the lean and yaw
reversals and advances the box by more than \SI{1}{m} in all five
trials, although every trial ends at the robot's base-tilt safety
trigger.

The results demonstrate repeatable short-horizon transfer of a learned
humanoid pivot gait without direct box-state sensing. The deployed
policy cannot sense or correct its drift relative to the box. Sustained
transport will therefore require estimating phase and heading drift from
onboard sensing and adding controlled stopping.
 
\ifanonymous\else
\section*{Acknowledgment}
This work was partially supported by JST Moonshot R\&D Program Japan Grant Number
JPMJMS263D.
\fi

\bibliographystyle{IEEEtran}
\bibliography{references}

\begin{thebibliography}{10}
\providecommand{\url}[1]{#1}
\csname url@samestyle\endcsname
\providecommand{\newblock}{\relax}
\providecommand{\bibinfo}[2]{#2}
\providecommand{\BIBentrySTDinterwordspacing}{\spaceskip=0pt\relax}
\providecommand{\BIBentryALTinterwordstretchfactor}{4}
\providecommand{\BIBentryALTinterwordspacing}{\spaceskip=\fontdimen2\font plus
\BIBentryALTinterwordstretchfactor\fontdimen3\font minus
  \fontdimen4\font\relax}
\providecommand{\BIBforeignlanguage}[2]{{%
\expandafter\ifx\csname l@#1\endcsname\relax
\typeout{** WARNING: IEEEtran.bst: No hyphenation pattern has been}%
\typeout{** loaded for the language `#1'. Using the pattern for}%
\typeout{** the default language instead.}%
\else
\language=\csname l@#1\endcsname
\fi
#2}}
\providecommand{\BIBdecl}{\relax}
\BIBdecl

\bibitem{hwangbo2019learning}
J.~Hwangbo \emph{et~al.}, ``Learning agile and dynamic motor skills for legged
  robots,'' \emph{Science Robotics}, vol.~4, no.~26, 2019, art. no. eaau5872.

\bibitem{lee2020learning}
J.~Lee, J.~Hwangbo, L.~Wellhausen, V.~Koltun, and M.~Hutter, ``Learning
  quadrupedal locomotion over challenging terrain,'' \emph{Science Robotics},
  vol.~5, no.~47, 2020, art. no. eabc5986.

\bibitem{rudin2022learning}
N.~Rudin, D.~Hoeller, P.~Reist, and M.~Hutter, ``Learning to walk in minutes
  using massively parallel deep reinforcement learning,'' in \emph{Proc. 5th
  Conf. Robot Learning (CoRL)}, 2022, pp. 91--100.

\bibitem{siekmann2021periodic}
J.~Siekmann, Y.~Godse, A.~Fern, and J.~Hurst, ``Sim-to-real learning of all
  common bipedal gaits via periodic reward composition,'' in \emph{Proc. IEEE
  Int. Conf. Robot. Autom. (ICRA)}, 2021, pp. 7309--7315.

\bibitem{leziart2026cat}
P.-A. L{\'e}ziart, M.~Morisawa, and F.~Kanehiro, ``Learning humanoid
  loco-manipulation with constraints as terminations,'' in \emph{Proc.
  IEEE/SICE Int. Symp. Syst. Integr. (SII)}, 2026, pp. 325--332.

\bibitem{zheng2025embracing}
C.~Zheng \emph{et~al.}, ``Embracing bulky objects with humanoid robots:
  Whole-body manipulation with reinforcement learning,'' \emph{arXiv preprint
  arXiv:2509.13534}, 2025.

\bibitem{pan2026unwieldy}
Z.~Pan, Y.~Yu, and Z.~Zou, ``Learning humanoid loco-manipulation for
  transporting unwieldy construction objects,'' in \emph{Proc. 43rd Int. Symp.
  Autom. Robot. Construction (ISARC)}, 2026, pp. 9--16.

\bibitem{aiyama1993pivoting}
Y.~Aiyama, M.~Inaba, and H.~Inoue, ``Pivoting: A new method of graspless
  manipulation of object by robot fingers,'' in \emph{Proc. IEEE/RSJ Int. Conf.
  Intell. Robots Syst. (IROS)}, 1993, pp. 136--143.

\bibitem{mason1986mechanics}
M.~T. Mason, ``Mechanics and planning of manipulator pushing operations,''
  \emph{Int. J. Robot. Res.}, vol.~5, no.~3, pp. 53--71, 1986.

\bibitem{harada2003pushing}
K.~Harada, S.~Kajita, K.~Kaneko, and H.~Hirukawa, ``Pushing manipulation by
  humanoid considering two-kinds of {ZMPs},'' in \emph{Proc. IEEE Int. Conf.
  Robot. Autom. (ICRA)}, 2003, pp. 1627--1632.

\bibitem{murooka2015pushing}
M.~Murooka, S.~Nozawa, Y.~Kakiuchi, K.~Okada, and M.~Inaba, ``Whole-body
  pushing manipulation with contact posture planning of large and heavy object
  for humanoid robot,'' in \emph{Proc. IEEE Int. Conf. Robot. Autom. (ICRA)},
  2015, pp. 5682--5689.

\bibitem{hou2018fast}
Y.~Hou, Z.~Jia, and M.~T. Mason, ``Fast planning for 3d any-pose-reorienting
  using pivoting,'' in \emph{Proc. IEEE Int. Conf. Robot. Autom. (ICRA)}, 2018,
  pp. 1631--1638.

\bibitem{yoshida2010pivoting}
E.~Yoshida \emph{et~al.}, ``Pivoting based manipulation by a humanoid robot,''
  \emph{Autonomous Robots}, vol.~28, no.~1, pp. 77--88, 2010.

\bibitem{yoshida2010planning}
------, ``Planning whole-body humanoid locomotion, reaching, and
  manipulation,'' in \emph{Motion Planning for Humanoid Robots}.\hskip 1em plus
  0.5em minus 0.4em\relax Springer, 2010, pp. 99--128.

\bibitem{murooka2014pivoting}
M.~Murooka, S.~Noda, S.~Nozawa, Y.~Kakiuchi, K.~Okada, and M.~Inaba,
  ``Achievement of pivoting large and heavy objects by life-sized humanoid
  robot based on online estimation control method of object state and
  manipulation force,'' \emph{J. Robot. Soc. Japan}, vol.~32, no.~7, pp.
  595--602, 2014.

\bibitem{zhang2021graph}
A.~Zhang, K.~Koyama, W.~Wan, and K.~Harada, ``Controlling pivoting gait using
  graph model predictive control,'' \emph{IEEE Access}, vol.~9, pp.
  73\,757--73\,770, 2021.

\bibitem{antonova2017pivoting}
R.~Antonova, S.~Cruciani, C.~Smith, and D.~Kragic, ``Reinforcement learning for
  pivoting task,'' \emph{arXiv preprint arXiv:1703.00472}, 2017.

\bibitem{heess2017emergence}
N.~Heess \emph{et~al.}, ``Emergence of locomotion behaviours in rich
  environments,'' \emph{arXiv preprint arXiv:1707.02286}, 2017.

\bibitem{andrychowicz2020learning}
M.~Andrychowicz \emph{et~al.}, ``Learning dexterous in-hand manipulation,''
  \emph{Int. J. Robot. Res.}, vol.~39, no.~1, pp. 3--20, 2020.

\bibitem{ng1999policy}
A.~Y. Ng, D.~Harada, and S.~Russell, ``Policy invariance under reward
  transformations: Theory and application to reward shaping,'' in \emph{Proc.
  Int. Conf. Mach. Learn. (ICML)}, 1999, pp. 278--287.

\bibitem{westervelt2004experimental}
E.~R. Westervelt, G.~Buche, and J.~W. Grizzle, ``Experimental validation of a
  framework for the design of controllers that induce stable walking in planar
  bipeds,'' \emph{Int. J. Robot. Res.}, vol.~23, no.~6, pp. 559--582, 2004.

\bibitem{unitree2024g1}
{Unitree Robotics}, ``G1 humanoid robot -- product specifications,''
  \url{https://www.unitree.com/g1/}, 2024, accessed 2026-07-20.

\end{thebibliography}

\end{document}